\documentclass[sigconf]{acmart}
\usepackage{graphicx}
\usepackage{subcaption} 
\usepackage{multirow}
\usepackage{multicol}
\usepackage{color}
\usepackage{newfloat}
\usepackage{textcomp}
\usepackage{xcolor}

\def\BibTeX{{\rm B\kern-.05em{\sc i\kern-.025em b}\kern-.08em
     T\kern-.1667em\lower.7ex\hbox{E}\kern-.125emX}}
     
\usepackage{booktabs}       
\usepackage{nicefrac}       
\usepackage{microtype}      
\usepackage{wrapfig}
\usepackage{algorithm}
\usepackage{algpseudocode}

\usepackage{geometry} 
\newcommand{\ParFor}[1]{%
    \textbf{Parallel for:} #1
}
\newcommand{\EndParFor}{%
    \textbf{End Parallel for}
}

\AtBeginDocument{%
  \providecommand\BibTeX{{%
    \normalfont B\kern-0.5em{\scshape i\kern-0.25em b}\kern-0.8em\TeX}}}

\setcopyright{acmlicensed}
\copyrightyear{2024}
\acmYear{2024}
\acmDOI{XXXXXXX.XXXXXXX}

\acmConference[Preprint '24]{Irregular Applications}{Nov, 2024}{Atlanta, GA}
\acmISBN{978-1-4503-XXXX-X/18/06}

\begin{document}

\title{Efficient Parallel Multi-Hop Reasoning: A Scalable Approach for Knowledge Graph Analysis}

\author{Jesmin Jahan Tithi}
\affiliation{%
  \institution{Parallel Computing Labs, Intel Corporation}
  \city{Santa Clara}
  \state{CA}
  \country{USA}
  \postcode{95054}
}
\email{jesmin.jahan.tithi@intel.com}

\author{Fabio Checconi}
\affiliation{%
  \institution{Parallel Computing Labs, Intel Corporation}
  \city{Santa Clara}
  \state{CA}
  \country{USA}
  \postcode{95054}
}
\email{fabio.checconi@intel.com}

\author{Fabrizio Petrini}
\affiliation{%
  \institution{Parallel Computing Labs, Intel Corporation}
  \city{Santa Clara}
  \state{CA}
  \country{USA}
  \postcode{95054}
}
\email{fabrizio.petrini@intel.com}
\renewcommand{\shortauthors}{J.J. Tithi, et al.}

\begin{abstract}
Multi-hop reasoning (MHR) is a process in artificial intelligence and natural language processing where a system needs to make multiple inferential steps to arrive at a conclusion or answer. In the context of knowledge graphs or databases, it involves traversing multiple linked entities and relationships to understand complex queries or perform tasks requiring a deeper understanding. Multi-hop reasoning is a critical function in various applications, including question answering, knowledge base completion, and link prediction. It has garnered significant interest in artificial intelligence, machine learning, and graph analytics. 

This paper focuses on optimizing MHR for time efficiency on large-scale graphs, diverging from the traditional emphasis on accuracy which is an orthogonal goal. We introduce a novel parallel algorithm that harnesses domain-specific learned embeddings to efficiently identify the top K paths between vertices in a knowledge graph to find the best answers to a three-hop query. Our contributions are: (1) We present a new parallel algorithm to enhance MHR performance, scalability and efficiency. (2) We demonstrate the algorithm's superior performance on leading-edge Intel and AMD architectures through empirical results.

We showcase the algorithm's practicality through a case study on identifying academic affiliations of potential Turing Award laureates in Deep Learning, highlighting its capability to handle intricate entity relationships. This demonstrates the potential of our approach to enabling high-performance MHR, useful to navigate the growing complexity of modern knowledge graphs.

\end{abstract}


\maketitle

\section{Introduction}
Multi-hop reasoning (MHR) is a process in artificial intelligence and natural language processing where a system needs to make multiple inferential steps to arrive at a conclusion or answer. In the context of knowledge graphs or databases, it involves traversing multiple linked entities and relationships to understand complex queries or perform tasks that require a deeper level of understanding.

For example, in a question-answering system, a single-hop query might ask, "Where was Albert Einstein born?" The answer can be found directly by looking up the birthplace of Albert Einstein in a knowledge graph. In contrast, a multi-hop query might be, "Which university did the physicist who proposed the theory of relativity work for?" To answer this, the system must first identify Albert Einstein as the physicist in question, then find the theory of relativity associated with him, and finally, determine the university where he worked.

Multi-hop reasoning is crucial for tasks that require connecting disparate pieces of information that are not immediately adjacent in a knowledge graph. It enables systems to handle more complex queries and provides a framework for more sophisticated and contextually aware AI applications. 

Encyclopedic KGs like Wikidata serve as repositories of structured world knowledge, connecting entities through relational triples (head, relation, tail) that are instrumental for downstream knowledge-driven applications. Despite their breadth, these KGs are incomplete, often lacking critical relational data. MHR emerges as a solution to infer these missing links, thereby enriching the KG and enhancing its utility for applications that demand a comprehensive knowledge base. 

The essence of MHR lies in its ability to traverse expansive networks, extracting salient insights by exploring multiple intermediate vertices or paths. The MHR problem can be framed as the task of finding the "best" $K$ paths between two vertices, $s$ and $t$, within a graph $G$ where vertices and edges could be of different types and the edge weights can be computed based on some learned embeddings \cite{guo2016entity} of the vertices and edge types reflecting domain-specific knowledge, guide the selection of paths that reveal subtle and often hidden relationships. These embeddings can be learned from a prior ML model such as a Graph Neural Network \cite{gilmer2020message,khemani2024review} or BERT \cite{reimers2019sentence} training. 

While existing research predominantly concentrates on enhancing the accuracy of MHR processes, we assert the equal importance of performance optimization. An unoptimized algorithm can be significantly slower, which is untenable for time-sensitive applications such as question answering systems in Healthcare Decision Support Systems, Financial Fraud Detection, Real-Time Recommendation Engines, Interactive Voice Assistants, Emergency Response Systems, High-Frequency Trading (HFT), etc.

Obtaining high performance on MHR is challenging since it involves large graph or database traversals and hashtable accesses which requires extensive random memory accesses and global synchronizations, which can pose significant barriers to achieving high performance. Scaling MHR is also challenging due to the large size of the input graphs and the associated learned embedding data. 

Embedding vector tables can reach terabyte-scale dimensions, which could exceed the memory capacity of a single server node. 

\begin{figure}[h]
\centering
\includegraphics[width=0.8\linewidth]{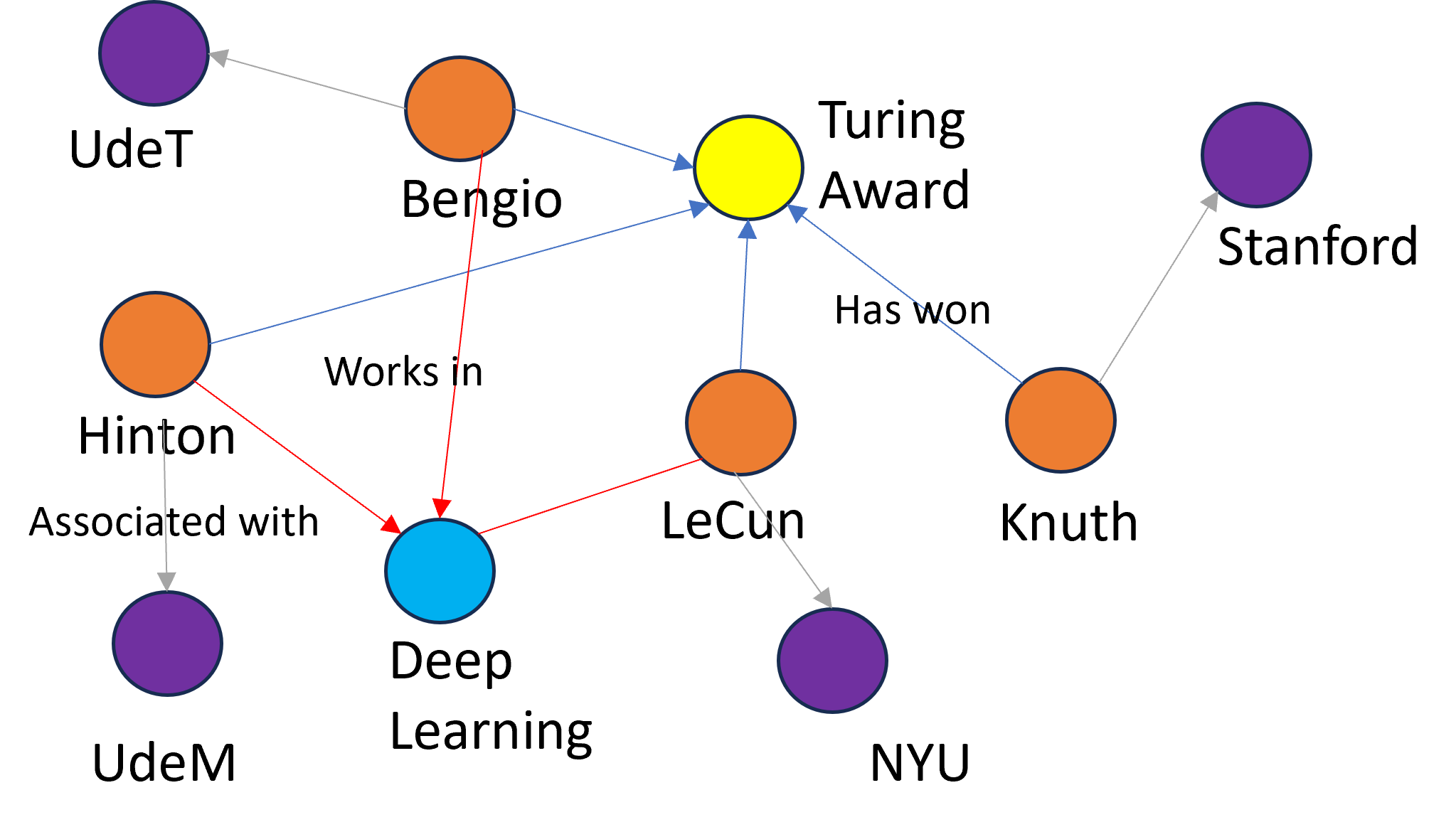} 
\\
\includegraphics[width=0.6\linewidth]{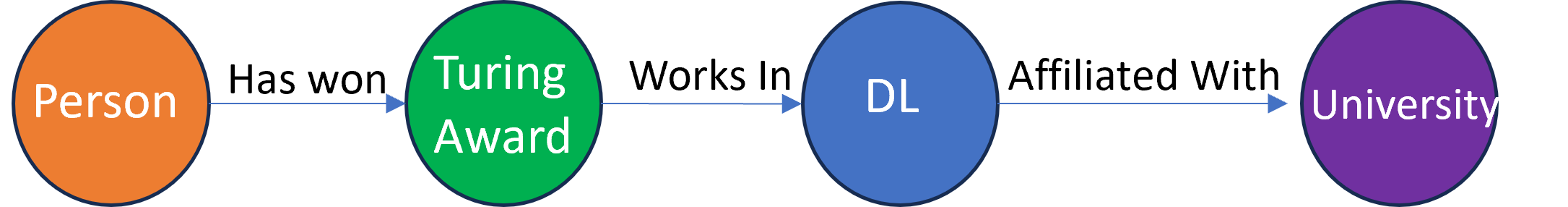} 
\caption{An example snippet of Knowledge Graph (KG) and the composition of a multi-hop reasoning query on the KG.}
\label{fig:CLQA}
\end{figure}

A practical example of the MHR problem is the identification of academic affiliations of potential Turing Award laureates within the Deep Learning (DL) discipline (see Figure \ref{fig:CLQA}). This problem is an example of complex logical query answering in a neural graph database and can be mapped to a three-hop reasoning problem as discussed later in the paper. We introduce an efficient parallel algorithm to address this MHR task. Our proposed model transcends academic applications, providing valuable insights into complex networks where MHR can uncover missing links, patterns, and predictions of substantial importance. 

This work presents an efficient parallel algorithm that speeds up the MHR problem. We begin with a basic parallel algorithm that a data scientist would typically use to solve the problem. We explain the potential performance improvement ideas and then propose an optimized algorithm to solve the same problem significantly faster compared to the basic one on any machine. Our approach uses concurrent hashmaps, k-heaps, and tree-based reduction to merge k-heaps to improve computational throughput. The primary contribution of our work is this optimized algorithm.

The contributions of this paper are threefolds:

\begin{itemize}
\item We frame the MHR problem as determining the best $K$ paths between two vertices in a graph, informed by a learned model. Given an input graph $G$ with entities $h, r$ connected with different types of edges/relationships $r$, and given the learned embedding vectors for all the entities and relationships in the graph, we try to answer, what are the $K$ best paths from $h$ to $r$. This framework can be generalized to diverse domain-specific applications, can be used to answer questions or suggest missing links in knowledge graphs.

\item We present a simple parallel algorithm for MHR that would be easy to implement but may not be performant. Then introduce an improved algorithm that utilizes concurrent hash tables, thread private k-heaps and tree merging of those heaps that would result in significantly faster solutions.

\item We show performance analysis of these algorithms on cutting-edge AMD EPYC and Intel SPR architectures, showcasing the practical effectiveness of our method.

\end{itemize}
Through these contributions, we aim to set a new precedent for performance-oriented MHR, enabling faster and more efficient reasoning over complex knowledge graphs.

\section{Related Work}

Multi-hop reasoning (MHR) is a concept in natural language processing (NLP) and knowledge graph (KG) analysis that involves answering complex queries by traversing multiple connected entities in a sequence. Formally, multi-hop queries, also known as path queries, are conjunctive queries (CQ) that form a linear chain. In such a chain, the tail of one projection serves as the head of the subsequent projection. Path queries can be resolved iteratively by retrieving the neighbors of the current node. Multi-hop queries can also be extended to include negation, allowing for more complex query formulations \cite{CLQAinNGR}.

MHR approaches often employ iterative methods like depth-first search (DFS) or breadth-first search (BFS) to navigate paths within a knowledge graph (KG) \cite{lao2010relational}. These methods may not consider the inherent meaning of the entity and relationships and regardless can become computationally intensive as the size of $K$ and the complexity of KGs increase.

Graph neural networks (GNNs) \cite{gilmer2020message,khemani2024review} have revolutionized learning over graph-structured data, offering sophisticated means to deduce relationships. These networks can be trained to generate entity and relationship embeddings \cite{guo2016entity}, which are dense vector representations of categorical variables in a reduced-dimensional space, encapsulating the relationships between categories.

TransE \cite{bordes2013translating} embedding model posits that the embedding of a tail entity should be near the sum of the head entity and relation embeddings if the triplet (head, relation, tail) is valid. A number of query reasoning approaches have been developed to address multi-hop path queries through the use of sequence models that leverage chainable knowledge graph embeddings, such as TransE as demonstrated by Guu et al. \cite{guu2015traversing}, or employ recurrent neural networks like LSTM, as utilized by Das et al. \cite{das2016chains}. These efforts represent some of the initial attempts to apply embeddings and neural network methodologies to solve multi-hop path queries.
 
The Hop, Union, Generate (HUG) \cite{zhao-etal-2023-hop} framework introduces an explainable MHR framework that uses a pre-trained model to score paths for their relevance to the reasoning task. The Tree-of-Mixed-Thought (ToMT) \cite{hu2023treeofmixedthought} model combines rapid, one-stop reasoning with iterative refinement from a learned model, striving for a balance between efficiency and accuracy in MHR tasks.

None of the prior work is focused on improving runtime performance of MHR on an HPC system. Our work introduces an efficient parallel algorithm for MHR. It concurrently traverses the relavant knowledge graph nodes, using TransE-style relations to find the most promising next paths to follow. Our algorithm accelerates the reasoning process and scales well with the number of cores on modern servers.

\section{Proposed Algorithm}
In this section, we discuss the problem and the algorithm, and provide steps to optimize the algorithm on a shared memory system. 
\subsection{The Problem and Its Mapping to MHR}
Given a knowledge graph, we are trying to answer the following question: who are the top K persons that are Winners of Turing Award, and also work in the deep learning field and then find the top K universities these top K persons are likely to be affiliated with. This problem can be rephrased as: 1) find top K $X$ entities, who are most likely to be related to $A$ with a relation $R1$, 2) out of those Top K, find those who are likely to be related to $B$ with relation $R2$ and 3) then find top K $C$s entities that are related to those top $X$ with some other relationship $R3$. We can solve this problem using a three-hop reasoning process - a particular type of Multi-hop Reasoning (MHR). Note that this process can be extended to other problems without losing generality.

For this particular problem, the first hop involves identifying nodes in the knowledge graph that are award winners and has a connection to the "TURING AWARD" entity. We score these nodes based on the TransE relation (h + r - t), where "h" represents the TURING AWARD embedding, "r" represents the award-winning relationship, and "t" represents the embedding for each person on the list. Based on these scores, we select the top individuals who are likely to receive the Turing Award.

In the second hop, we determine which of these individuals work in the field of deep learning. We use the TransE relationship again, where "h" represents "DEEP LEARNING", "r" represents the "works in" relationship, and "t" represents the person of interest. We rank them based on the (h + r - t) TransE score.

In the final hop, we aim to determine the most probable university affiliations of these individuals. In this case, we use the "affiliated with" relationship, where "h" represents the person, and "t" represents a university.

Our problem requires three different inputs. Firstly, we have a knowledge graph represented as edge lists in the format (h, r, t). Here, "h" represents the source node, "t" represents the destination node, and "r" represents the relationship between them. These nodes represent different entities each with their pre-learned embedding vectors. Secondly, we have a relationship embedding table that contains learned meanings of each relationship from a machine learning model. Finally, we have an embedding table that contains embedding vectors for each entity, which were also pre-learned using a machine-learning model.

\subsection{Simple Algorithm}
Listing \ref{algo:naive} shows a simple algorithm to solve the multi-hop reasoning problem to discover potential academic affiliations for Turing Award prospects in the field of deep learning, utilizing pre-learned entity and relation embeddings. Here are the steps in the algorithm:

{\bf{Initialization of Graph Data:}} The algorithm begins by parsing triples (head entity $h$, relation $r$, tail entity $t$) from an edge list file $E$. These triples represent the edges of the knowledge graph and are used to initialize the graph structure. Edge tables $E_r$ are populated in parallel for each relation $r$, indexing the head entities $h$ as keys. This step is synchronized using locks to prevent concurrent write conflicts.

{\bf{Loading Embeddings:}} Entity embeddings are loaded from a file V into an entity embedding hash table $V_e$, with each entity uniquely identified by its ID. This operation is also performed in parallel with locks ensuring data integrity. Similarly, relation embeddings are loaded from a file R into a relation embedding hash table $R_e$, indexed by the relation type.

{\bf{Extraction of Entities and Universities:}} Person entities are extracted from the edge table $E_{AWARD WINNER}$ and stored in a hashtable or array P. This is done in parallel using locks. A set of universities U is extracted from the edge table $E_{AFFILIATED WITH}$ in parallel, forming the university table.

{\bf{Scoring and Ranking Process:}} A parameter K is set to define the number of top paths to be considered. A ranked list of persons $L_p$ is initialized as empty. In parallel, each person p related to the "Turing Award" is scored using the TransE model, and the pair (person p, score $S_p$) is added to the list $L_p$ atomically. A TopK heap $H_p$ is created from the list $L_p$ based on the scores, and the heap is sorted in descending order to produce a ranked list of persons $TP$.

{\bf{Deep Learning Association:}} For each top-ranked person ($p_k$, $S_{p_k}$) in $TP$, a score $S_{p_k, DL}$ is computed for their association with "deep learning," and the pair is added to the list $L_p$ using locks. A TopK heap $H_p$ is created from the list $L_p$ based on the scores, and the heap is sorted in descending order in $TP$. 

{\bf{University Affiliation Scoring:}} The process is repeated for each person in $TP$ to compute affiliation scores $S_{p_k, u}$ for each university $u$ in the set U. These scores are added to the university list $L_u$ using locks. A TopK heap $H_u$ is created from the list $L_u$ using the affiliation scores, and the heap is sorted in descending order to rank the universities. The algorithm concludes by returning a list of potential affiliations A for the top-scored individuals, representing the final output of the process.

\begin{algorithm*}
\caption{Parallel Multi-Hop Reasoning for Academic Affiliation Discovery}
\begin{algorithmic}[1]

\Require Edge list $E$, entity embeddings $V$, relation embeddings $R$
\Ensure Top academic affiliations $A$ for Turing Award prospects in deep learning

\State \textbf{Parallel} Parse $(h, r, t)$ from $E$ to initialize graph grouped by relation types $r$ \Comment{Initialize graph}
\State \textbf{Parallel} Populate $E_r$ using $h$ as key with locks \Comment{Populate edge tables}

\State \textbf{Parallel} Populate $V_e$ from $V$ using entity ID as key with locks \Comment{Load entity embeddings}
\State \textbf{Parallel} Populate $R_e$ from $R$ using relation type as key with locks \Comment{Load relation embeddings}

\State \textbf{Parallel} Extract $P$ from $E_{\text{AWARD WINNER}}$ with locks \Comment{Extract persons}
\State \textbf{Parallel} Extract $U$ from $E_{\text{AFFILIATED WITH}}$ with locks \Comment{Extract universities}

\State Set $K$ for the number of top paths \Comment{Define top paths}
\State Initialize $L_p \gets \emptyset$ \Comment{Initialize person list}

\State \textbf{Parallel} \ForAll{$p \in P$}
    \State Compute $S_p$ using TransE model
    \State Atomically insert $(p, S_p)$ to $L_p$
\EndFor
\State Create TopK heap $H_p$ from $L_p$ and sort it by score in $TP$\Comment{Rank persons}

\State \textbf{Parallel} \ForAll{$(p_k, S_{p_k}) \in TP$}
    \State Compute $S_{p_k, \text{DL}}$ for "deep learning" association
    \State Add $(p_k, S_{p_k})$ to $L_p$ with locks
\EndFor
\State Create TopK heap $H_p$ from $L_p$ and sort it by score in $TP$\Comment{Rank persons}

\ForAll{$(p_k, S_{p_k}) \in TP$}
   \State \textbf{Parallel} \ForAll{$u \in U$}
        \State Compute $S_{p_k, u}$ for "AFFILIATED WITH"
        \State Add $(u, S_{p_k, u})$ to $L_u$ with locks
    \EndFor
    \State Create TopK heap $H_u$ from $L_u$ and sort it by score in $A$\Comment{Rank universities}
    \State Print $A$ \Comment{Output affiliations}
\EndFor

\end{algorithmic}
\label{algo:naive}
\end{algorithm*}

This could be a standard algorithm to solve the multi hop reasoning problem in Python or C++. Although it is a good starting point, the algorithm has several inadequacies and scalability issues. In this paper, we introduce an improved algorithm that solves the same problem that can lead to a significantly lower runtime cost.

\subsection{Algorithmic Optimizations}

For performance enhancement, we introduce a custom thread-safe hashtable tailored for high-concurrency insertions. This approach mitigates the inefficiencies inherent in standard STL hashtables that rely on user-level locking. Our specialized hashtable is designed to handle the storage demands of embedding vectors, which may scale to terabytes, thereby reducing synchronization overhead and increasing the throughput of our MHR algorithm. While hardware-accelerated hashing presents a further opportunity for optimization, it is beyond the scope of this paper.

In the context of scoring and ranking individuals, we address the issue of thread conflicts in shared hashmaps by utilizing thread-private k-heaps for storing (person, score) pairs. These k-heaps, being priority queues, maintain only the top K scores, offering several advantages:

\begin{itemize} \item Synchronization across threads during insertions is eliminated, enhancing scalability. \item Storage complexity is reduced from O(Persons) to O(threads * K), which is more practical for large datasets. \item K-heaps typically reside in thread-local storage, such as cache, leading to faster insertions compared to global hashmaps that require locks and are stored in main memory. \item By tracking only the top K elements per thread, we conserve memory and simplify the identification of the global top K elements in subsequent steps. \end{itemize}

\begin{algorithm*}
\begin{algorithmic}
\Function{ReduceTopKTree}{$localTopK$, $globalTopK$, $numThreads$, $threadId$}
    \State {\bf {Parallel}} \For{$stride \gets 1$ \textbf{to} $numThreads$ \textbf{by} $stride \times 2$}
        \If{$threadId$ is multiple of $2 \times stride$ \textbf{and} $threadId + stride < numThreads$}
            \State $otherThreadId \gets threadId + stride$
            \State Merge heaps $localTopK[threadId]$ and $localTopK[otherThreadId]$
        \EndIf
        \State Perform barrier synchronization
    \EndFor
    \State \textbf{single thread execution:}
    \State Update $globalTopK$ with $localTopK[0]$
\EndFunction
\end{algorithmic}
\caption{Tree-based reduction Algorithm to merge K-heaps from all threads into one single K-heap.}
\label{algo:ReduceTopKTree}
\end{algorithm*}

Upon completion of individual scoring by all threads, we merge the thread-local k-heaps into a unified k-heap to determine the global top K individuals. For this purpose, we use a tree-based heap merging algorithm, ReduceTopKTree, which employs a binary tree structure for efficient heap consolidation. The pseudocode for this algorithm is detailed in Algorithm~\ref{algo:ReduceTopKTree}. This can be further improved by using a quaternary tree merging. Note that, one could use locks to merge the K-heaps. However, based on our experiences, we anticipate that such approch would be less scalable on parallel machines with many threads.

The computeScorePerPerson algorithm, detailed in Algorithm \ref{algo:computeScorePerPerson}, systematically identifies individuals most likely associated with the Turing Award by leveraging their entity and relation embeddings. The process unfolds as follows:

Initially, the algorithm retrieves the Turing Award's entity and relation embeddings from their respective tables. These are then combined into a composite embedding, encapsulating the award's characteristics within the knowledge graph.

A normalization constant, gamma, is set to 1.0, and the algorithm is configured for parallel execution by initializing a local top-k heap for each thread. This setup facilitates the simultaneous processing of individual scores.

During the parallel computation phase, threads independently compute similarity scores for individuals by contrasting their embeddings with the composite Turing Award embedding. The score reflects the degree of association with the award, calculated as the normalized difference between the embeddings.

Each thread maintains a local top-k heap to store the highest scores, ensuring that only the most pertinent candidates are considered. Upon completion, these local heaps are consolidated into a global top-k heap through the reduction function descrived before.

The outcome is a global top-k heap that ranks individuals according to their association with the Turing Award, as determined by their embeddings. This parallel algorithm enables scalable and efficient identification of the most relevant candidates.

\begin{algorithm*}
\caption{Compute Top-k Persons related to the Turing Award}
\begin{algorithmic}[1]
\Require $EntityEmbeddingTable$, $RelationEmbeddingTable$
\Ensure $top\_k\_elements$ (MaxHeap)

\Function{computeScorePerPerson}{$EntityEmbeddingTable$, $RelationEmbeddingTable$, $top\_k\_elements$}
    \State $turing\_emb\_ptr \gets EntityEmbeddingTable.find(TURING\_AWARD)$
    \State $TURING\_AWARD\_EMB \gets turing\_emb\_ptr.embedding$
    \State $AWARD\_WINNER\_TABLE\_EMB \gets RelationEmbeddingTable[AWARD\_WINNER\_TABLE\_IDX].embedding$
    \State $turing\_arr \gets$ \Call{embeddingAggregation}{$TURING\_AWARD\_EMB$, $AWARD\_WINNER\_TABLE\_EMB$}
    \State $gamma \gets 1.0$
    \State $num\_persons \gets$ size of $person\_ids$
    \State $localTopK[num\_threads]$ \Comment{Array of MaxHeaps, one per thread}
    \ParFor{$id \gets 0$ to $num\_threads - 1$}
        \State $localTopK\_t \gets localTopK[id]$
        \For{$i \gets 0$ to $num\_persons - 1$}
            \State $person\_id \gets person\_ids[i]$
            \State $record \gets EntityScoreInfo(person\_id, -\infty)$
            \State $e, emb \gets EntityEmbeddingTable.find(person\_id)$
            \If{$e \neq HashmapStatus.Fail$}
                \State $total\_score \gets 0.0$
                \For{$j \gets 0$ to $EMBEDDING\_DIMENSION - 1$}
                    \State $total\_score \gets total\_score + |\text{turing\_arr[j]} - emb.embedding[j]|$
                \EndFor
                \State $record.score \gets gamma - total\_score$
            \EndIf
            \State $localTopK\_t.push(record)$
            \If{$localTopK\_t.size() > TOP\_K$}
                \State $localTopK\_t.pop()$
            \EndIf
        \EndFor
        \State \Call{reduceTopKTree}{$localTopK$, $top\_k\_elements$, $num\_threads$, $id$}
    \EndParFor
\EndFunction
\end{algorithmic}
\label{algo:computeScorePerPerson}
\end{algorithm*}

The next step in the process is to rank or rearrange these top-K Turing Award winning people based on their probability of working in the Deep Learning field. The computeScoreBasedOnWorksInDL function calculates a score for individuals potentially associated with the Turing Award and active in the Deep Learning field. It accepts entity and relation embeddings, a max heap for top-k elements, and identifiers for the head entity, relation, and individual.

The algorithm retrieves the relevant embeddings for the head entity and relation, then computes the L1 norm as the sum of absolute differences between the head entity plus relation embeddings and the individual's embedding. This score reflects the individual's alignment with the characteristics of the Turing Award and their contributions to Deep Learning.

An adjustment is made to the score by subtracting it from a constant, gamma. The resulting TransE score is used to rank the individual's relevance. The function ensures thread-safe operations while updating the max heap with the individual's score, maintaining only the top-k scores. This process identifies the most relevant candidates for the award within the Deep Learning context.

Again, in this case, the simple algorithm might use a global hashmap to keep the key (person) and score (association probability based on TransE) and instead, we use K-heaps.

The next step of the MHR process is to find the most likely university affiliations of these top-k people found in the prior step. The function computeAffiliationScore calculates the top-k affiliations for a given a person, based on their association with various universities. The function takes the entity and relation embedding tables, a max heap to store the top-k elements, and identifiers for the head entity and the relation of interest.

First, the function retrieves the embeddings for the head entity (person) and the specified relation (AFFILIATED\_WITH). These embeddings are aggregated to form a composite embedding array, affiliated\_arr, which represents the combined characteristics of the head entity and the relation.

The algorithm sets a constant, gamma, to 1.0, which will be used to adjust the final scores. In a parallelized block, each thread processes a subset of universities. The threads each maintain a local max heap, localTopK\_t, to store the top-k affiliation scores for the universities they process. For each university, the function computes a score by summing the absolute differences between the composite embedding array and the university's embedding. This score is adjusted by subtracting it from gamma, yielding the final affiliation score for the university.

The local top-k heaps are managed to ensure they contain the top-k scores. If a heap exceeds the top-k limit, the lowest scoring element is removed. After processing, the local heaps are merged into the global top-k heap using a reduction function, reduceTopKTree.

The final output is the global top-k heap, which contains the top-k university affiliations for the head entity based on their embeddings. This heap represents the most relevant affiliations, identifying the universities most closely associated with the person of interest. Algorithm \ref{algo:computeAffiliationScore} shows a pseudocode for the process. 

\begin{algorithm*}
\caption{Compute Top-k Affiliations for a Person}
\begin{algorithmic}[1]
\Require $EntityEmbeddingTable$, $RelationEmbeddingTable$, $headId$, $relationId$
\Ensure $top\_k\_elements$ (MaxHeap)

\Function{computeAffiliationScore}{$EntityEmbeddingTable$, $RelationEmbeddingTable$, $top\_k\_elements$, $headId$, $relationId$}
    \State $head\_emb\_ptr \gets EntityEmbeddingTable.find(headId)$
    \State $head\_emb \gets head\_emb\_ptr.embedding$
    \State $relation\_emb \gets RelationEmbeddingTable[relationId].embedding$
    \State $affiliated\_arr \gets$ \Call{embeddingAggregation}{$head\_emb$, $relation\_emb$}
    \State $gamma \gets 1.0$
    \State $num\_universitys \gets$ size of $university\_ids$
    \State $localTopK[num\_threads]$ \Comment{Array of MaxHeaps, one per thread}
    \ParFor{$id \gets 0$ to $num\_threads - 1$}
        \State $localTopK\_t \gets localTopK[id]$
        \For{$i \gets 0$ to $num\_universitys - 1$}
            \State $university\_id \gets university\_ids[i]$
            \State $record \gets EntityScoreInfo(university\_id, -\infty)$
            \State $e, uni\_emb\_ptr \gets EntityEmbeddingTable.find(university\_id)$
            \If{$e \neq HashmapStatus.Fail$}
                \State $total\_score \gets 0.0$
                \For{$j \gets 0$ to $EMBEDDING\_DIMENSION - 1$}
                    \State $total\_score \gets total\_score + |\text{affiliated\_arr[j]} - uni\_emb\_ptr.embedding[j]|$
                \EndFor
                \State $record.score \gets gamma - total\_score$
            \EndIf
            \State $localTopK\_t.push(record)$
            \If{$localTopK\_t.size() > TOP\_K$}
                \State $localTopK\_t.pop()$
            \EndIf
        \EndFor
        \State \Call{reduceTopKTree}{$localTopK$, $top\_k\_elements$, $num\_threads$, $id$}
    \EndParFor
\EndFunction
\end{algorithmic}
\label {algo:computeAffiliationScore}
\end{algorithm*}

\section{Performance Result}
\subsection{Dataset}
To demonstrate the performance improvement of our new algorithm, we used the WikiKG90Mv2\cite{hu2021ogblsc} dataset from the Stanford Open Graph Benchmark (OGB) dataset (ogb.stanford.edu). The WikiKG90Mv2 dataset is a subset of Wikidata, aimed at enhancing the graph's coverage by predicting missing triples. It contains 91,230,610 entities, 1,387 relations, and 601,062,811 triples (h, r, t), extracted from Wikidata dumps as of May 17th, June 7th, and June 28th, 2021, for training, validation, and testing. Each entity and relation is accompanied by text features, including titles and descriptions, with MPNet\cite{song2020mpnet} sentence embeddings \cite{reimers2019sentence} provided to capture semantic meanings. This dataset is utilized to assess the performance of our multi-hop reasoning algorithm. However, the dataset has been trimmed (embedding size = 8)  so that we can fit them in a single machine's memory.

The process of constructing a knowledge graph begins with reading an WikiKG90Mv2 edgelist file, where each line represents a triple consisting of a head, a relation, and a tail. These triples are then sorted into 1,387 distinct relation tables, with each (head, tail) pair being placed in the table corresponding to its relation. Following this, two separate files containing embeddings for approximately 91 million entities and all relations are read to construct the entity and relation embedding tables, respectively.

A specialized table is created to store person entities, initially populated by examining the "AWARD WINNER" relation table. Similarly, a university/institution table is assembled, primarily using the "AFFILIATED WITH" relation table. Table \ref{tab:datastats} shows different statistics about the data.

\begin{table}[htbp]
  \centering
  \caption{Property of the Dataset used to test MHR}
    \begin{tabular}{lrr}
    \toprule
    \multicolumn{3}{p{17em}}{\bf Parameter for Edge Data} \\
    \midrule
    Num\_edges & 6.01E+08 &  \\
    File size (B) & 1.44E+11 &  \\
    Record size (B) & 256   &  \\
    LINE SIZE & 256   &  \\
     \midrule
    \multicolumn{3}{p{17em}}{\bf Parameter for EntityEmbedding} \\
     \midrule
    \multicolumn{1}{p{9.715em}}{num\_entity} & 91000000 &  \\
    \multicolumn{1}{p{9.715em}}{File Size} & 4.168E+12 &  \\
    \multicolumn{1}{p{9.715em}}{Record Size} & 3072  &  \\
    EMD size & 768   &  \\
    LINE SIZE & 3072  &  \\
     \midrule
    \multicolumn{3}{p{17em}}{\bf Parameter for RelationEmbedding} \\
     \midrule
    Num\_relations & 1387  &  \\
    File Size & 6.48E+10 &  \\
    Record Size & 3072  &  \\
    EMD size & 768   &  \\
    LINE SIZE & 3072  &  \\
     \midrule
    \multicolumn{3}{p{17em}}{\bf Parameter for PersonTable} \\
     \midrule
    Num\_person & 539603 &  \\
    person\_record\_size & 16    &  \\
           \midrule
    \multicolumn{3}{p{17em}}{\bf Parameter for UniversityTable} \\
     \midrule
    Num\_university & 89689 &  \\
    University record size & 16    &  \\
           \midrule
   \multicolumn{3}{p{17em}}{\bf Parameter for TopK} \\
     \midrule
    Top\_k & 50    &  \\
           \midrule
    \multicolumn{3}{p{17em}}{\bf Multihop Reasoning} \\
     \midrule
    EMB Used & 8   &  \\
    Elements & 6.92E+08 &  \\
    \bottomrule
    \end{tabular}%
  \label{tab:datastats}%
\end{table}%
\subsection{Systems Specifications}
Table \ref{tab:sysspec} presents the system parameters of the Intel and AMD platforms we used to evaluate the performance of our algorithm implementations. Both systems have similar capabilities, but the EPYC system has a higher average clock speed and slightly higher core counts while having slightly less bandwidth (about 30\%). Additionally, the EPYC system has 8 NUMA dies, whereas SPR has 2. Based on these specifications, we expect SPR to be faster for the bandwidth-bound portions of the code, while EPYC could better in handling latency due to its more cores. However, the EPYC system might be slightly more affected by NUMAness than SPR. Overall, we anticipate that these platforms will perform similarly. The EPYC system belongs to the Perlmutter Supercomputing cluster hosted at Berkeley, while SPR is an in-house system. We collected the bandwidth information using Memory Latency Checker which runs on Intel systems but does not run on EPYC. For EPYC, we used the stream triad benchmark to measure the bandwidth. 

We utilized OpenMP and C/C++ languages to develop parallel implementations for both the simple and the optimized algorithms. Afterwards, we compiled the program using g++ compiler while applying standard optimization flags like "-O3 -fopenmp -Ofast" on both platforms.

\begin{table*}[tbhp]
  \centering
  \caption{Systems Specifications Used to Test The Algorithms}
    \begin{tabular}{lrr}
    \toprule
  
    \textbf{System Parameters} & \multicolumn{1}{l}{\textbf{SPR}} & \multicolumn{1}{l}{\textbf{EPYC}} \\
      \hline
    \#cores/sockets & 56    & 64 \\
      \hline
    Model & Intel(R) Xeon(R) Platinum 8480L & AMD EPYC 7763 64-Core Processor \\
    \hline
    \#threads & 112   & 128 \\
    \hline
    \#hyperthreads & 224   & 256 \\
    \hline
    \#sockets & 2     & 2 \\
    \hline
    Clock Speed (GHz) & 2     & 2.45 \\
    \hline
    Memory Size (GB) & 527 total, 257 per socket & 527 total, 65 per socket  \\
    \hline
    \#Memory Nodes/Dies & 2     & 8 \\
    \hline
    Total Memory BW (GB/s) & 527 (all read), 474 (stream triad) & 379 (stream triad) \\
    \hline
    Cache Size & 107 MB/core & 512 KB/core \\
    \hline
    Compiler Version (g++) & 11.4.1 20230605 & 12.3.0 \\
    \bottomrule
    \end{tabular}%
  \label{tab:sysspec}%
\end{table*}%

\subsection{Performance Trends}
In this section, we discuss the performance results of the MHR code on the SPR platform and compare it with the optimized version's performance on both SPR and EPYC platforms. We utilized the command "KMP\_AFFINITY=compact OMP\_PLACES=cores numactl -a -b -i 0" to affinitize the threads to cores, which helped in creating a balanced and interleaved memory allocation across the memory nodes. This setting was crucial for achieving satisfactory performance on the EPYC machine. 

\begin{figure}[htbp]
\centering
\includegraphics[width=0.8\linewidth]{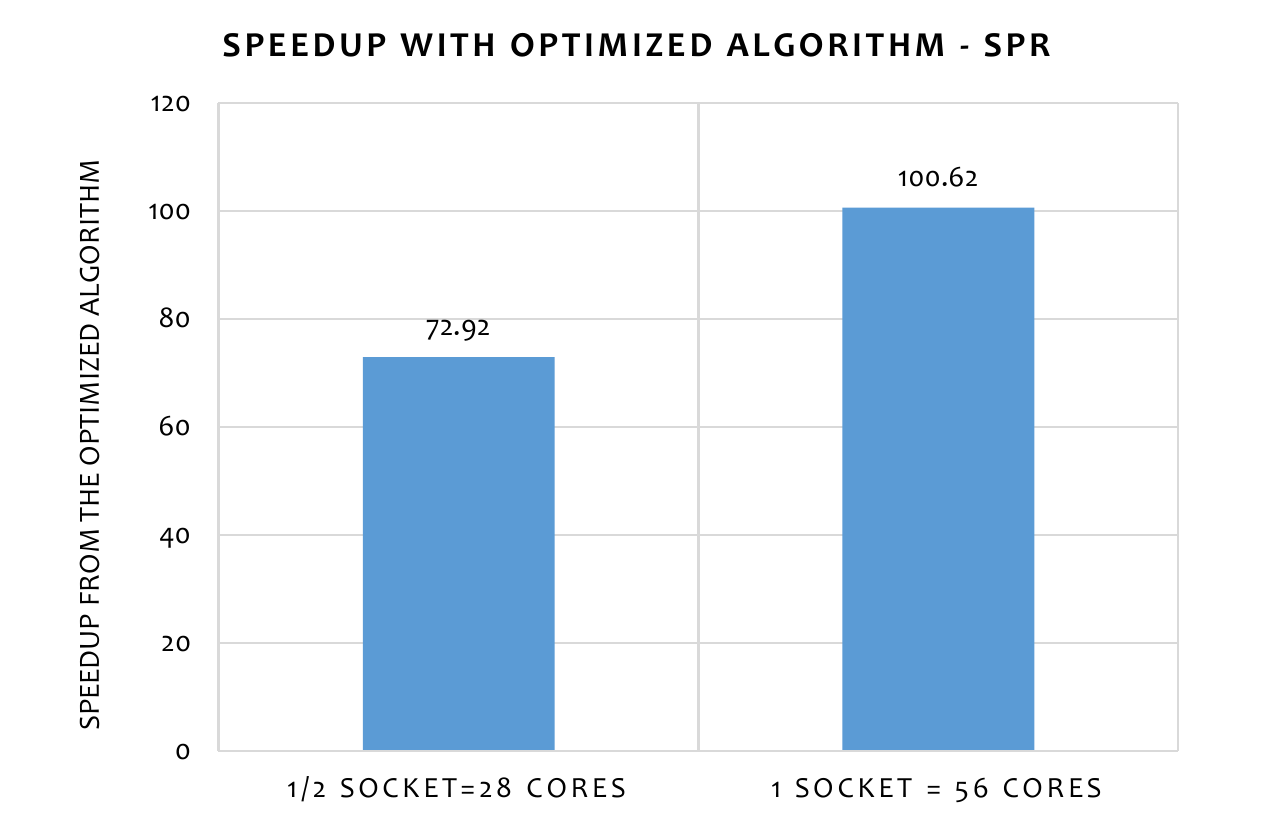} 
\caption{Performance Benefit of the old algorithm compared on SPR Platform.}
\label{fig:spr_old_new}
\end{figure}

Figure \ref{fig:spr_old_new} illustrates the difference in the performance of the simple algorithm and the optimized algorithm running on a single socket of the SPR node with 56 cores. The results show that the optimized algorithm is 100 times faster than the simple algorithm when executed on a single socket of the SPR node. The speedup difference between 28 cores and 56 cores indicates that the optimized algorithm is more scalable than the simple algorithm. This is because the speedup increases as the number of cores increases. The optimized algorithm uses a customized hashtable with fine-grained locking and private K-heaps instead of a global hashtable, as explained in the previous sections all of which contribute to such a speedup.

\begin{figure}[thbp]
\centering
\begin{subfigure}[b]{0.494\linewidth}
\includegraphics[width=\linewidth]{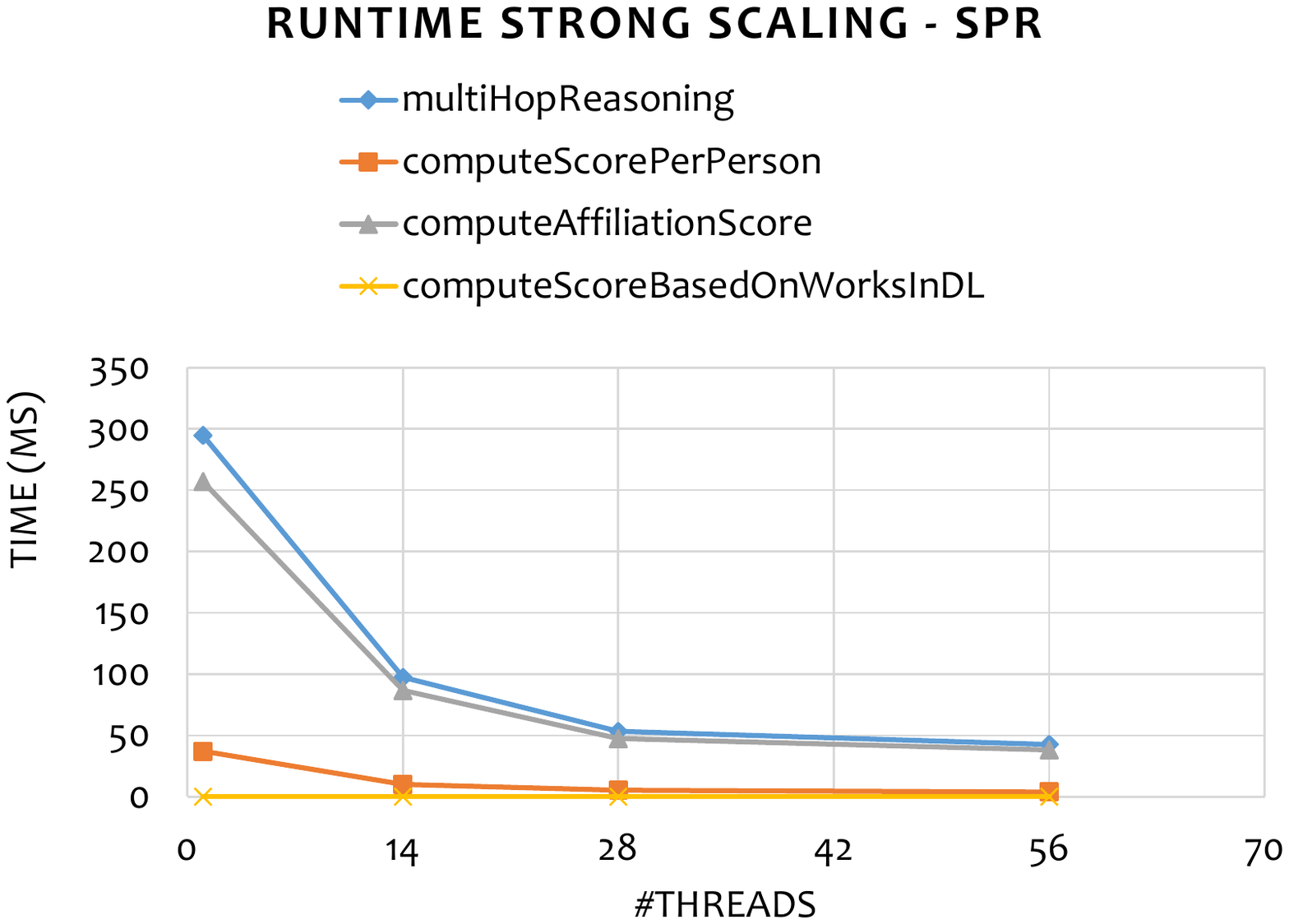}
\caption{Runtime scaling}
\label{fig:spr_strong_scaling}
\end{subfigure}
\begin{subfigure}[b]{0.494\linewidth}
\includegraphics[width=\linewidth]{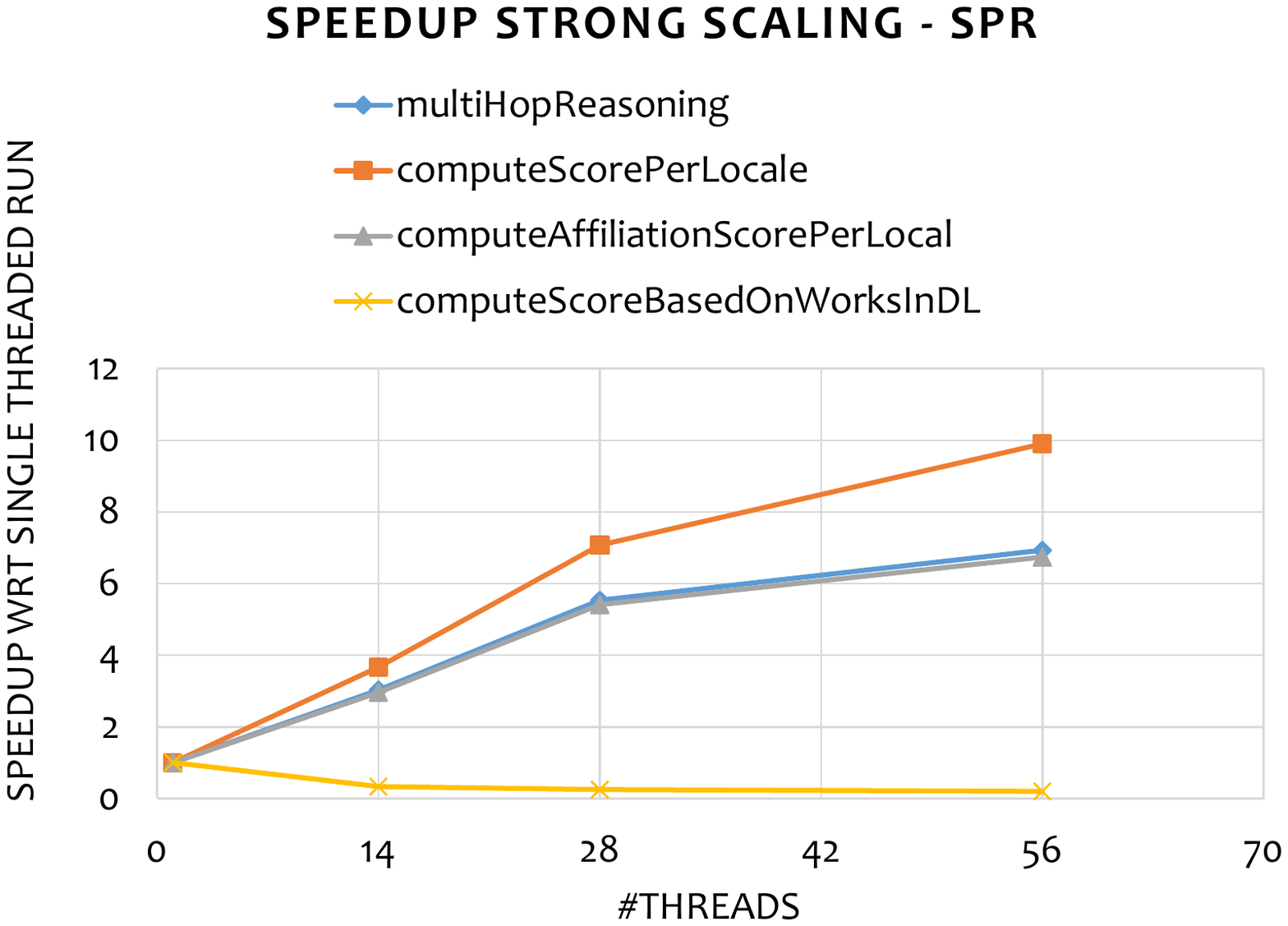}
\caption{Speedup scaling}
\label{fig:spr_strong_scaling_speed}
\end{subfigure}
\caption{Strong Scaling of Runtime of the MHR Algorithm on SPR Platform.}
\label{fig:spr_strong_runtime}
\end{figure}

The graph depicted in Figure \ref{fig:spr_strong_runtime} demonstrates the strong scaling of an optimized algorithm on the SPR platform. The multHopReasoning task represents the entire process, while computeScorePerPerson, computeScoreBasedOnWorksInDL, and computeAffiliationScore are the three sub-tasks of multHopReasoning, as described in the algorithm description. One important point to note is that among these three sub-tasks, computeAffiliationScore takes the most time since it is executed O(Persons * Universities) times, whereas the other two sub-tasks are executed O(Persons) and O(top-K) times, respectively. The graph in Figure \ref{fig:spr_strong_runtime} shows that the algorithm performs well at different core counts and scales well on 56 cores of one socket of SPR. The algorithm's performance improves by approximately 7 times on 56 cores. It is worth mentioning that the MHR algorithm is very sparse and involves frequent hash lookup mixed with embedding gathering, making it a challenging workload to obtain high performance.

\begin{figure}[htbp]
\centering
\begin{subfigure}[b]{0.494\linewidth}
\includegraphics[width=\linewidth]{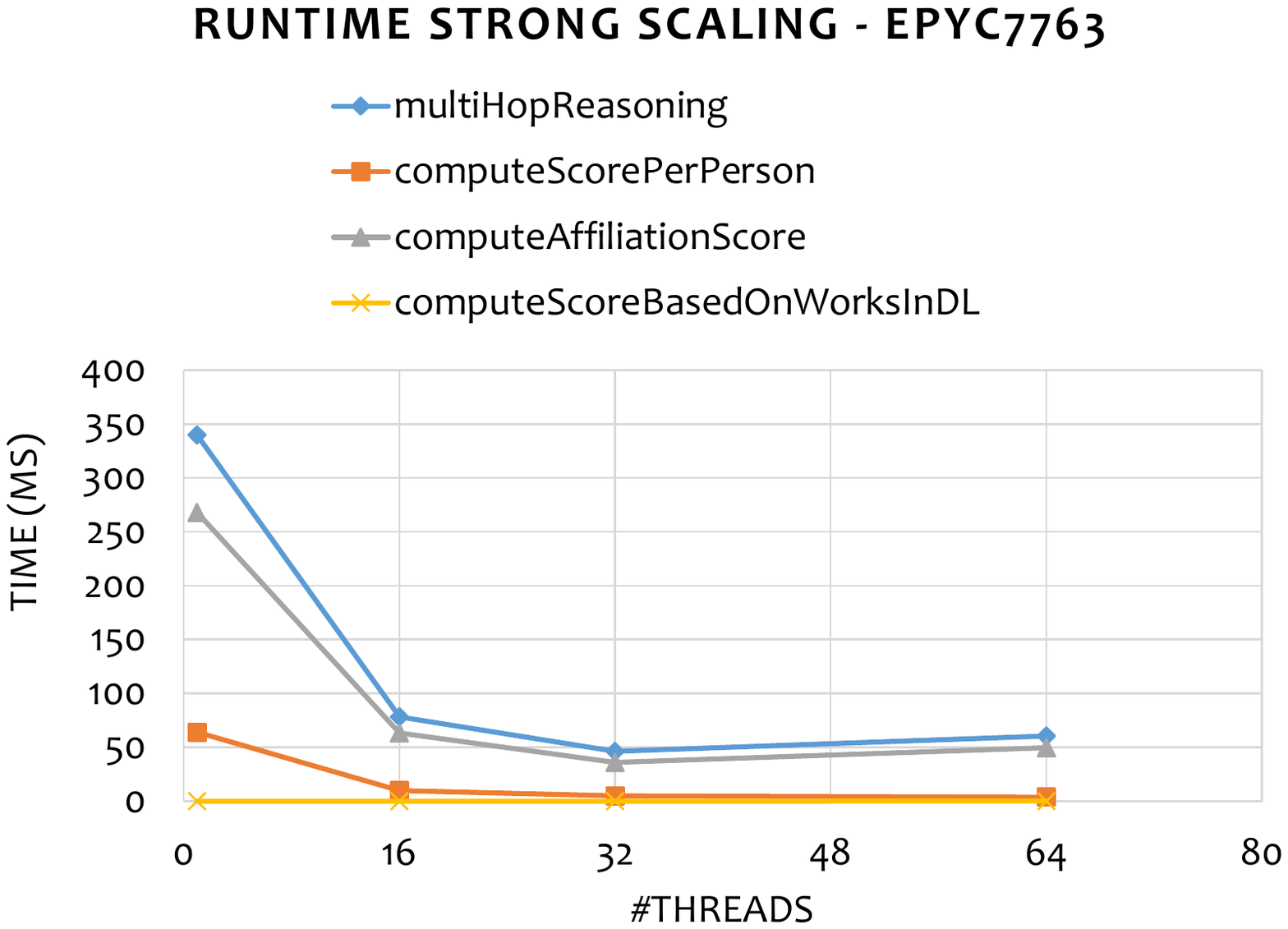}
\caption{Runtime scaling}
\label{fig:spr_strong_scaling}
\end{subfigure}
\begin{subfigure}[b]{0.494\linewidth}
\includegraphics[width=\linewidth]{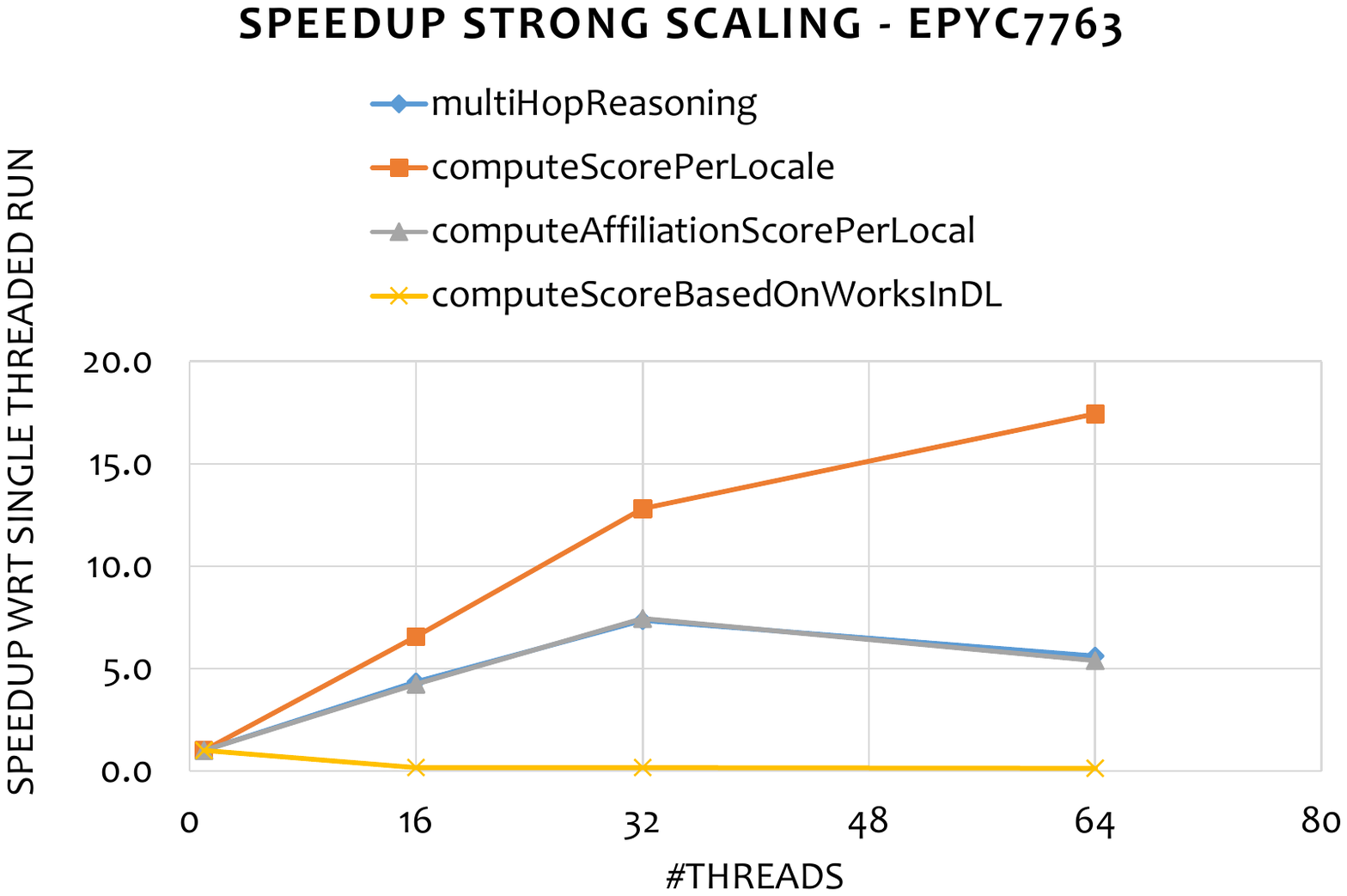}
\caption{Speedup scaling}
\label{fig:spr_strong_scaling_speed}
\end{subfigure}
\caption{Strong Scaling of Runtime of the MHR Algorithm on EPYC Platform.}
\label{fig:epyc_strong_runtime}
\end{figure}

Figure \ref{fig:epyc_strong_runtime} illustrates the optimized algorithm's performance on the EPYC platform across varying core counts. The algorithm exhibits robust scaling up to the utilization of one socket, equivalent to four NUMA nodes. However, beyond three NUMA nodes, no further performance gains are noted. Notably, the algorithm achieves a 5.6x speedup on 64 cores and a 7.2x speedup on 32 cores. Critical to achieving this scaling is the use of thread pinning and interleaved memory allocation, as specified by "OMP\_PLACES=cores numactl -a -b -I 0-3". 
Without such measures, performance is markedly reduced, underscoring the importance of data interleaving in mitigating NUMA-related slowdowns. Future work will focus on addressing NUMA latencies to further enhance algorithm performance on NUMA architectures.

\begin{figure}[tbph]
\centering
\includegraphics[width=0.8\linewidth]{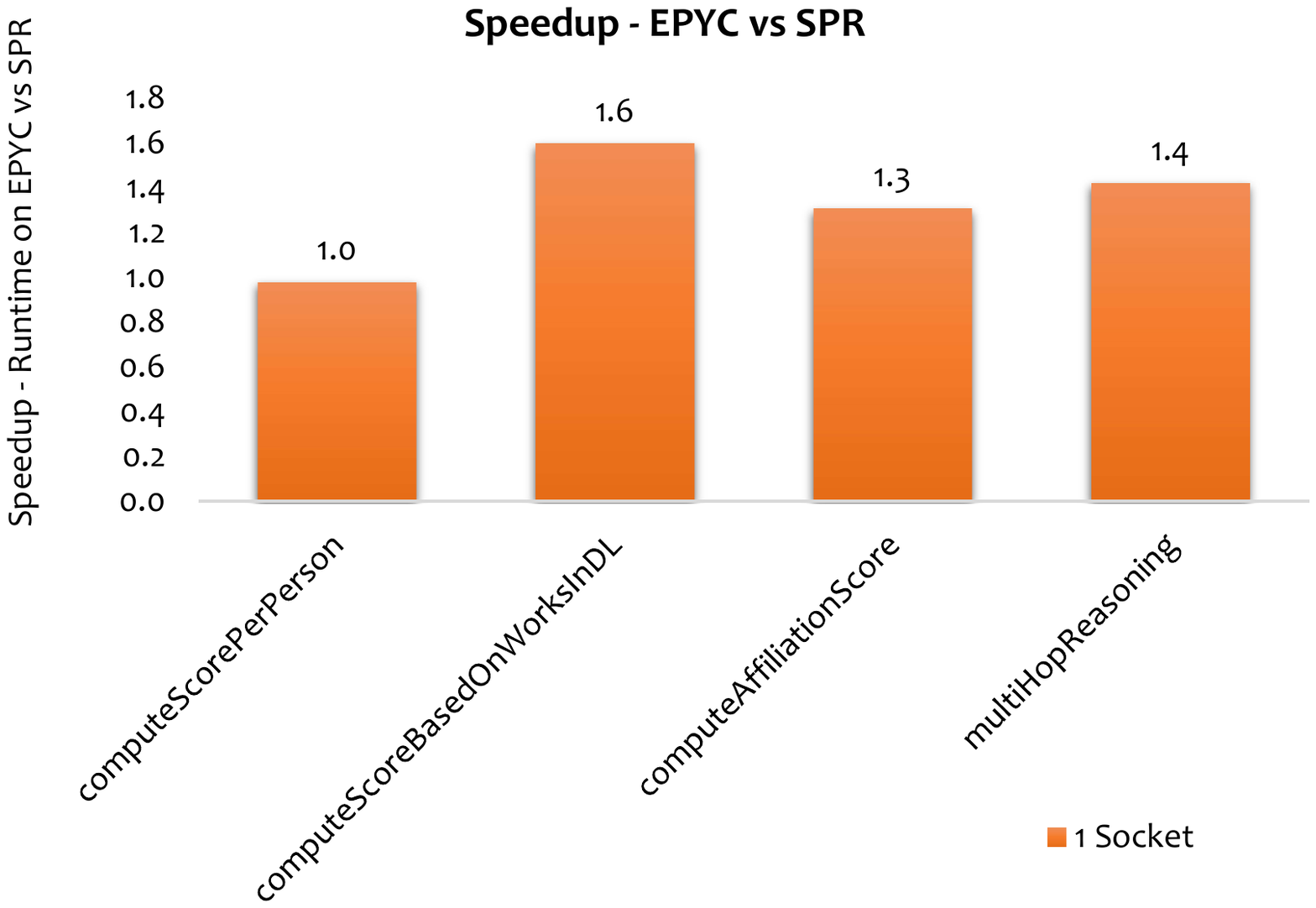} 
\caption{Performance difference between SPR and EPYC.}
\label{fig:epyc_spr}
\end{figure}

In evaluating the performance disparity between SPR and EPYC platforms, Figure \ref{fig:epyc_spr} reveals that SPR surpasses EPYC by 40\% when operating on a single socket configuration. This performance advantage is primarily due to SPR's superior memory bandwidth and EPYC's fasters clock speed. The algorithm experiences a greater impact from the Non-Uniform Memory Access (NUMA) effect on the EPYC system compared to SPR, indicating that SPR's architecture mitigates NUMA-related performance degradation more effectively.

Finally, we also experimented with a version of our optimized algorithm that uses locks to merge in the global top-K heap instead of using the tree-based reduction and as expected this version performed slightly worse than the tree merge algorithm.
\section{Generalization of the MHR Algorithm}

The versatility of our MHR algorithm extends beyond the specific use case of identifying academic affiliations for Turing Award candidates. Its underlying structure is designed to generalize across various domains where multi-hop queries are essential. By abstracting the core components of the algorithm, we can apply it to different types of entities and relationships within any knowledge graph.

The generalization process involves redefining the entities and relationships of interest and then applying the same reasoning steps. For instance, the algorithm can be adapted to identify connections between medical conditions and treatments, or to trace supply chain networks from manufacturers to end consumers.

Below is a high-level pseudocode that illustrates the generalization of the algorithm.
\begin{algorithm*}
\caption{Generalized Three-Hop Reasoning Algorithm}
\begin{algorithmic}[1]
\scriptsize
\Require GeneralEntityEmbeddingTable, GeneralRelationEmbeddingTable
\Ensure GeneralTopKResults (MaxHeap)

\Function{GeneralizedMultiHopReasoning}{GeneralEntityEmbeddingTable, GeneralRelationEmbeddingTable, GeneralTopKResults}
    \State Define general entities and relationships of interest
    \State GeneralEntityEmbeddingTable $\gets$ LoadEntityEmbeddings()
    \State GeneralRelationEmbeddingTable $\gets$ LoadRelationEmbeddings()
    
    \State FirstHopResults $\gets$ computeRelation1ScorePerEntity1(GeneralEntityEmbeddingTable, GeneralRelationEmbeddingTable, GeneralTopKResults)
    \State SecondHopResults $\gets$ computeScoreBasedOnRelation2Entity2(FirstHopResults, GeneralTopKResults)
    \State FinalResults $\gets$ computeRelation3ScorePerEntity1(SecondHopResults, GeneralTopKResults)
    
    \State GeneralTopKResults $\gets$ MergeTopKHeaps(FinalResults)
    \State \Return GeneralTopKResults
\EndFunction
\end{algorithmic}
\label{algo:GeneralizedMHR}
\end{algorithm*}
In this pseudocode, LoadEntityEmbeddings() and LoadRelationEmbeddings() are placeholders for functions that would load the relevant embeddings based on the domain-specific problem. The computeRelation1ScorePerEntity1(maps to computeScorePerPerson), computeScoreBasedOnRelation2Entity2\\(maps to computeScoreBasedOnWorksInDL), and computeRelation3ScorePerEntity1(maps to computeAffiliationScorePerLocal) functions are reused from the Turing Award problem, demonstrating the algorithm's adaptability. The MergeTopKHeaps function represents the final step of consolidating results from multiple threads.
\begin{algorithm}
\caption{Generic Multi-Hop Reasoning Algorithm}
\begin{algorithmic}[1]
\scriptsize
\Require Graph $G$, Embeddings $E$, Source $S$, Target $T$, Number of Hops $NUM\_HOPS$, Top $K$
\Ensure Results heap $results$

\Function{Multihop\_reasoning\_generic}{$G$, $E$, $S$, $T$, $NUM\_HOPS$, $topK$, $results$}
    \State $path \gets$ vector of int
    \State $path.insert(S)$ \Comment{Initialize path with source vertex}
    \State $total\_paths \gets (\Call{pow}{topK, NUM\_HOPS - 1} - 1) / (topK - 1)$
    \State Define $paths[total\_paths \times 2]$ \Comment{Array of paths}
    \State $curr, next \gets$ pointers to path arrays
    \State $curr\_size \gets 0$
    \State $next\_size \gets 0$
    \State $curr[curr\_size++] \gets path$ \Comment{Start with initial path}
    
    \For{$level \gets 0$ \textbf{to} $NUM\_HOPS - 1$}
        \State \#pragma omp parallel for num\_threads(min($curr\_size$, $total\_threads$))
        \For{$p \gets 0$ \textbf{to} $curr\_size - 1$}
            \State \Call{expand\_path}{$curr[p]$, $next$, $next\_size$, $G$, $E$, $T$, $topK$, $results$}
        \EndFor
        \State \Call{swap}{$curr$, $next$}
        \State $curr\_size \gets next\_size$
        \State $next\_size \gets 0$
    \EndFor
\EndFunction
\end{algorithmic}
\label{algo:MultihopReasoningGeneric}
\end{algorithm}

\begin{algorithm}
\caption{Expand Path in Multi-Hop Reasoning}
\begin{algorithmic}[1]
\scriptsize
\Require Current path $path$, Next paths array $next$, Size of next paths array $next\_size$, Graph $G$, Embeddings $E$, Target node $T$, Number of hops $NUM\_HOPS$, Top $K$ paths, Results heap $results$
\Ensure Updated next paths array $next$, Updated size $next\_size$, Updated results heap $results$

\Function{expand\_path}{$path$, $next$, $next\_size$, $G$, $E$, $T$, $NUM\_HOPS$, $topK$, $results$}
    \State $horizon\_node \gets path[\text{end}]$
    \State $neighbors \gets G[horizon\_node].neighbors$
    \If{$neighbors$ is not empty}
        \State $path\_emb \gets$ EmbeddingForPath($path$, $E$)
        \State $score\_heap \gets$ new max heap of size $topK$
        \State \bf {Parallel} \For{$i \gets 0$ to $|neighbors| - 1$}
            \State $neighbor \gets neighbors[i]$
            \If{not $neighbor$ in $path$}
                \State $extended\_path \gets path + [neighbor]$
                \State $extended\_emb \gets$ ExtendEmbedding($path\_emb$, $E[neighbor]$)
                \State $score \gets$ ComputeScore($extended\_emb$)
                \If{$neighbor = T$}
                    \State \bf {Atomically} $results$.insert($\{extended\_path, score\}$)
                \Else
                    \State \bf {Atomically} $score\_heap$.insert($\{extended\_path, score\}$)
                \EndIf
            \EndIf
        \EndFor
        \State \bf {Atomically} UpdateNextPaths($score\_heap$, $next$, $next\_size$, $topK$)
    \EndIf
\EndFunction
\end{algorithmic}
\label{algo:expandPath}
\end{algorithm}
Algorithms \ref{algo:MultihopReasoningGeneric}, \ref{algo:expandPath} show another example of generic multi-hop reasoning algorithm where the number of hops is also a parameter. The Multihop\_reasoning\_generic algorithm is designed to perform multi-hop reasoning on a given knowledge graph. It takes as input a graph $G$, a set of embeddings $E$, a source node $S$, a target node $T$, the number of hops $NUM\_HOPS$, and the number of top paths $K$ to find. The algorithm outputs a heap of results that contains the top paths from the source to the target.

The algorithm begins by initializing a path with the source node. It then calculates the total number of paths to explore based on the number of hops and the top $K$ paths. Two arrays of paths, curr and next, are defined to keep track of the current and next set of paths to explore. The algorithm starts with the initial path and iteratively expands each path in parallel using the expand\_path function.

The expand\_path function takes a path and expands it by exploring its neighboring nodes. For each neighbor, it checks if the neighbor is already in the path to avoid cycles. If the neighbor is not in the path, it is added, and the path embedding is extended. The score of the extended path is computed, and if the neighbor is the target node, the path and its score are added to the results heap. Otherwise, the path and its score are added to a local max heap of scores. After exploring all neighbors, the top paths from the local max heap are added to the next array of paths.

The algorithm continues this process for the specified number of hops, swapping the curr and next arrays at each iteration. The final results heap contains the top paths from the source to the target, as determined by the computed scores.

This generalized approach allows the MHR algorithm to be applied to a wide range of problems, making it a powerful tool for extracting insights from complex, interconnected data.

The three-hop reasoning problem presented in this paper is a particular instance of the generalized problem and the algorithmic optimizations presented for the three-hop problem are applicable to the general problem as well.
\section{Conclusions}
In this study, we have presented novel parallel algorithms designed to optimize multi-hop reasoning (MHR) tasks on large-scale knowledge graphs. Our approach has been tested using the WikiKG90Mv2 dataset, a subset of Wikidata, which serves as a challenging benchmark due to its extensive size and the rich semantic information it encapsulates. Despite the necessity to trim the dataset to fit within the memory constraints of a single machine, our algorithm demonstrated significant performance improvements.

We evaluated the effectiveness of the algorithm on two high-performance computing platforms, SPR and EPYC, each with unique architectural features. Our findings show that the optimized algorithm achieves significant speed improvements, surpassing the simple algorithm by a factor of 100 on a single socket in the SPR system. This increase in efficiency is due to the algorithm's ability to better manage access to the hashmap, resulting in fewer access conflicts, less synchronization overhead, and overall higher work efficiency compared to the simple algorithm due to the usage of private K-heaps and tree-based merging. 

On the EPYC platform, the algorithm exhibited strong scaling up to the limits of a single socket, with diminishing returns observed beyond 3 NUMA nodes. Thread pinning and balanced memory allocation were crucial in realizing the full potential of the EPYC system, underscoring the importance of memory access patterns in optimizing MHR tasks.

Our findings underscore the significance of memory bandwidth and the impact of NUMA effects on the performance of MHR algorithms. The SPR platform's architecture provided a notable advantage in memory-intensive operations, leading to a 40\% performance improvement over the EPYC system in single-socket configurations.

The insights gained from this research contribute to the broader understanding of MHR optimization strategies, highlighting the critical role of system architecture in achieving efficient parallel processing. This study contributes to the field of MHR by demonstrating the potential of algorithmic optimization to leverage high-performance computing architectures effectively. Our findings provide a foundation for future advancements in MHR algorithms, aiming for enhanced scalability and efficiency in processing complex knowledge graphs.

\section{Acknowledgement}
This research is based upon work supported by the Office of the Director of National Intelligence (ODNI), Intelligence Advanced Research Projects Activity (IARPA), through the Advanced Graphical Intelligence Logical Computing Environment (AGILE) research program, under Army Research Office (ARO) contract number <W911NF22C0081>. The views and conclusions contained herein are those of the authors and should not be interpreted as necessarily representing the official policies or endorsements, either expressed or implied, of the ODNI, IARPA, ARO, or the U.S. Government.

\bibliographystyle{ACM-Reference-Format}
\bibliography{sample-base}

\appendix

\end{document}